\documentclass{article}

\usepackage{times}
\usepackage{graphicx} 
\usepackage{subfigure} 
\usepackage{tikz}
\usepackage{natbib}

\usepackage{algorithm}
\usepackage{algorithmic}
\usepackage{amsmath}

\usepackage{amssymb}

\usepackage[accepted]{icml2017}

\icmltitlerunning{Multiscale Hierarchical Convolutional Networks}

\begin{document} 
\newcommand{\softmax}{\rm smax}
\newcommand{\rb}{\rangle}
\newcommand{\lb}{\langle}
\newcommand{\om}{\omega}
\newcommand{\la}{\lambda}
\newcommand{\Real} {{\rm Real}}
\newcommand{\R} {{\mathbb R}}
\newcommand{\E} {{\mathbb E}}
\newcommand{\N} {{\mathbb N}}
\newcommand {\h} {{\mathfrak h}}
\newcommand{\Z} {{\mathbb Z}}
\newcommand{\C} {{\mathbb C}}
\newcommand{\HH} {{\cal H}}
\newcommand{\Ld} {{\bf L^2}}

\definecolor{one}{HTML}{5497A7}
\definecolor{two}{HTML}{5FAD56}
\definecolor{three}{HTML}{F2C14E}
\definecolor{four}{HTML}{F78154}
\definecolor{five}{HTML}{B4436C}

\newtheorem{theorem}{Theorem}[section]
\newtheorem{definition}{Definition}[section]
\newtheorem{proposition}{Proposition}[section]
\newtheorem{corollary}{Corollary}[section]
\twocolumn[
\icmltitle{Multiscale Hierarchical Convolutional Networks}



\begin{icmlauthorlist}
\icmlauthor{J\"{o}rn-Henrik Jacobsen}{uva}
\icmlauthor{Edouard Oyallon}{ens} \\
\icmlauthor{St\'{e}phane Mallat}{ens} 
\icmlauthor{Arnold W.M. Smeulders}{uva}
\end{icmlauthorlist}
\icmlaffiliation{uva}{Informatics Institute, University of Amsterdam, Amsterdam}
\icmlaffiliation{ens}{D\'{e}partement Informatique, Ecole Normale Sup\'{e}rieure, Paris}
\icmlcorrespondingauthor{J\"{o}rn-Henrik Jacobsen}{j.jacobsen@uva.nl}
\icmlkeywords{deep learning, representation learning, convolutional neural network}

\vskip 0.3in
]



\printAffiliationsAndNotice{} 

\begin{abstract} 
Deep neural network algorithms are difficult to analyze because they lack structure
allowing to understand the properties of underlying transforms and invariants. 
Multiscale hierarchical convolutional networks are structured deep convolutional
networks where layers are indexed by progressively higher dimensional attributes, which
are learned from training data.
Each new layer is computed with multidimensional convolutions along spatial and
attribute variables.
We introduce an efficient implementation of such networks where the dimensionality
is progressively reduced by averaging intermediate
layers along attribute indices. Hierarchical networks are tested on CIFAR image data bases where they obtain comparable precisions to state of the art networks, with much fewer parameters. We study some properties of the attributes learned from these databases.
\end{abstract} 

\section{Introduction}
\label{intro}

Deep convolutional neural networks have demonstrated impressive performance for classification and regression tasks over a wide range of generic problems including images, audio signals, but also for game strategy, biological, medical, chemical and physics data \cite{lecun2015deep}. However, their mathematical properties remain mysterious and we are currently
not able to relate their performance to the properties of the classification problem.

Classifying signals in high dimension requires to eliminate non-informative variables, and hence contract and reduce space dimensionality in appropriate directions. Convolutional Neural Networks (CNN) discover these directions via backpropagation algorithms \citep{lecun1989backpropagation}. Several studies show numerically that linearization increases with depth \cite{zeiler2014visualizing}, but we do not know what type of information is preserved or eliminated. The variabilities which can be eliminated are mathematically defined as the
group of symmetries of the classification function \cite{mallat2016understanding}. It is 
the group of transformations (not necessarily linear) which preserves the labels of the classification problem. Translations usually belong to the symmetry group, and invariants to translations
are computed with spatial convolutions, followed by a final averaging. 
Understanding a deep neural network classifier requires specifying its symmetry group and invariants besides translations, especially of non-geometrical nature. 

To achieve this goal, we study highly structured
multiscale hierarchical convolution networks
introduced mathematically in \cite{mallat2016understanding}.
Hierarchical networks give explicit information on invariants 
by disentangling, progressively more signal attributes as the depth increases.
The deep network
operators are multidimensional convolutions along attribute indices. Invariants
are obtained by averaging network layers along these attributes. 
Such a deep network can thus
be interpreted as a non-linear mapping of the classification symmetry group into
a multidimensional translation group along attributes, which is 
commutative and hence flat.
Signal classes are mapped into manifolds which are 
progressively more flat as depth increases.

Section \ref{genericCNN}  reviews important properties of generic CNN architectures 
\cite{lecun2015deep}. Section \ref{MultiHiera} 
describes multiscale hierarchical convolutional 
networks, which are particular CNNs where 
linear operators are multidimensional convolutions along progressively more attributes.
Section \ref{fast} describes an efficient implementation, which reduces 
inner layers dimensions by computing invariants with an averaging along attributes. 
Numerical experiments on the CIFAR database show that this hierarchical network obtains
comparable performances to state of the art CNN architectures, with a reduced number
of parameters. Multiscale hierarchical network 
represent the symmetry group as multidimensional translations along non-linear
attributes. Section \ref{struct} studies the structuration obtained by
this deep network. This architecture provides a mathematical and experimental framework
to understand deep neural network classification properties.
The numerical results are reproducible and code is available online using TensorFlow and Keras\footnote{https://github.com/jhjacobsen/HierarchicalCNN}.

\begin{figure*}
\center
\includegraphics[width=12cm]{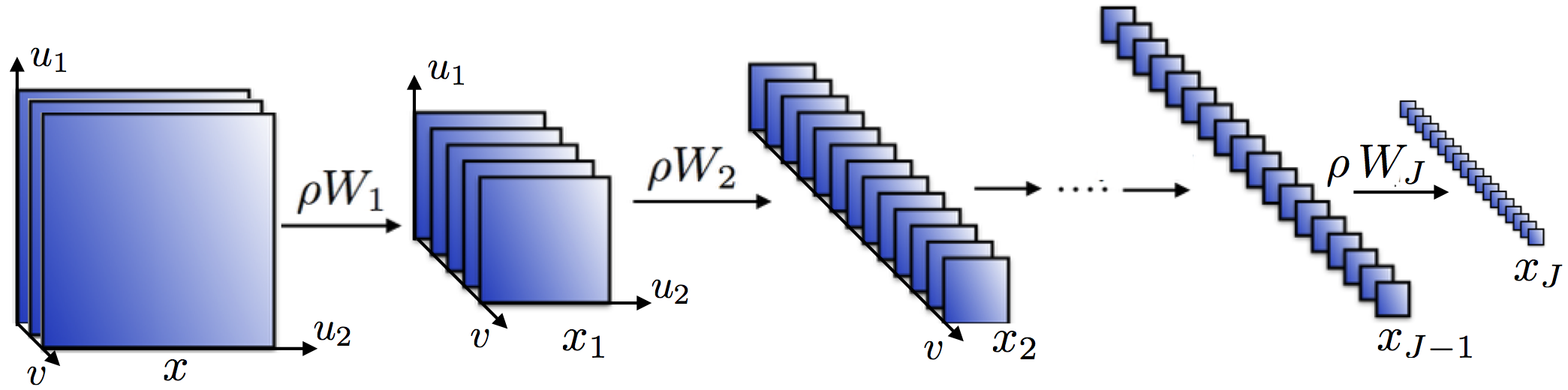}
\caption{A deep convolution network compute each layer $x_j$ with 
a cascade of linear operators $W_j$ and pointwise non-linearities $\rho$.}
\label{figure2}
\end{figure*}

\section{Deep Convolutional Networks and Group Invariants}
\label{genericCNN}
A classification problem associates 
a class $y= f(x)$ to any vector $x \in \R^N$ of $N$ parameters.
Deep convolutional networks transforms $x$ into multiple layers 
$x_j$ of coefficients at depths $j$, whose dimensions
are progressively reduced after a certain depth \cite{lecun2010convolutional}. 
We briefly review their properties.

We shall numerically
concentrate on color images $x(u,v)$ where $u = (u_1,u_2)$ are the
spatial coordinates and $1 \leq v \leq 3$ is the index of a color channel.
The input $x(u,v)$ may, however, correspond 
to any other type of signals. For sounds, 
$u = u_1$ is time and $v$ may be the index of audio channels
recorded at different spatial locations.

Each layer is an array of signals $x_j (u,v)$ where $u$ is the
native index of $x$, and $v$ is a $1$-dimensional 
channel parameter.
A deep convolutional network iteratively
computes $x_{j+1} = \rho \,W_{j+1}\, x_j$ with $x_0 = x$,
as illustrated in Figure \ref{figure2}.  
Each $W_{j+1}$ computes sums over $v$
of convolutions along $u$, with filters of small support. 
It usually also incorporates a batch
normalization \cite{ioffe2015batch}.
The resolution of 
$x_j (u,v)$ along $u$ is progressively reduced by a subsampling 
as $j$ increases until an averaging in the final output layer. 
The operator $\rho (z)$ is a pointwise non-linearity
In this work, we shall use exponential linear units
ELU \cite{clevert2015fast}. It transforms
each coefficient $z(t)$ plus a bias $c = z(t) + b$
into $c$ if $c < 0$ and $e^c - 1$ if $c < 0$.

As the depth increases,
the discriminative variations of $x$ along $u$
are progressively transferred to the
channel index $v$. At the last layer $x_J$, $v$ stands for the class index and
$u$ has disappeared. An estimation $\tilde y$ of the signal class $y = f (x)$ 
is computed by applying a soft-max to $x_J (v)$.
It is difficult to understand the meaning of this
channel index $v$ whose size and properties changes with depth. It 
mixes multiple unknown signal attributes with an arbitrary ordering.
Multiscale hierarchical convolution networks will
adress this issue by imposing a high-dimensional hierarchical structure
on $v$, with an ordering specified by the translation group. 

In standard CNN, each $x_j = \Phi_j x$ is computed 
with a cascade of convolutions and non-linearities 
\[\Phi_j = \rho\, W_j\,...\,\rho\, W_1 ,\]
whose supports along $u$ increase with the depth $j$.
These multiscale operators replace $x$ by the
variables $x_j$ to estimate the class $y = f(x)$.
To avoid errors, this change of variable must be discriminative, 
despite the dimensionality reduction, in the sense that
\begin{equation}
\label{discrimin}
\forall (x,x') \in \R^{2N}~~\Phi_j (x) = \Phi_j (x')~~\Rightarrow~~
f (x) = f (x')~.
\end{equation}
This is necessary and sufficient to guarantee that
there exists a classification function $f_j$ such that
$f = f_j \, \Phi_j$ and hence
\[
\forall x \in \R^N~~,~~f_j (x_j) = f(x).
\]
The function $f(x)$ can be characterized by its
groups of symmetries. A group of symmetries of $f$ 
is a group of operators $g$ which transforms any $x$
into $x' = g.x$ which belong to the same class:
$f(x) = f(g.x)$. The discriminative property (\ref{discrimin})
implies that if $\Phi_j (x) = \Phi_j (g.x)$ 
then $f(x) = f(g.x)$. The discrimination property (\ref{discrimin}) is thus
equivalent to impose that groups of symmetries 
of $\Phi_j$ are groups of symmetries of $f$. 
Learning appropriate change of variables can thus be interpreted as
learning progressively more symmetries of $f$ \cite{mallat2016understanding}.
The network must be sufficiently flexible to compute change of variables 
$\Phi_j$ whose symmetries approximate the symmetries of $f$.

Deep convolutional networks are cascading
convolutions along the spatial variable $u$ so that $\Phi_j$ is covariant to
spatial translations. If $x$ is translated along $u$ then $x_j = \Phi_j x$ 
is also translated along $u$. This covariance implies that for all $v$, 
$\sum_u x_j (u,v)$ is invariant to translations of $x$. 
Next section explains how to extend this property to higher dimensional
attributes with multidimensional convolutions. 

\section{Multiscale Hierachical Convolution Networks}
\label{MultiHiera}

Multiscale hierachical networks are highly structured convolutional
networks defined in \cite{mallat2016understanding}. The one-dimensional index $v$ is replaced
by a multidimensional vector of attributes $v = (v_1,...,v_j)$ and 
all linear operators $W_j$ are convolutions over $(u,v)$.
We explain their construction and a specific architecture
adapted to an efficient learning procedure. 
 
Each layer $x_j (u,v)$ is indexed by a vector of multidimensional 
parameters $v = (v_1,...,v_{j})$ of dimension $j$. 
Each $v_k$ is an ``attribute''  of $x$ which is learned 
to discriminate classes $y = f(x)$.
The operators $W_j$ are defined as convolutions along a group which is
a parallel transport
in the index space $(u,v)$. With no loss of generality, in this implementation, the transport is a multidimensional translation along $(u,v)$. 
The operators $W_j$ are therefore multidimensional convolutions, 
which are covariant to translations along $(u,v)$. As previously explained,
this covariance to translations implies that the sum
$\sum_{v_k} x_j (u,v_0,...,v_{j})$ is
invariant to translations of previous layers along $v_k$.
A convolution of $z(u,v)$ by a filter $w(u,v)$ of
support $S$ is written 
\begin{equation}
z \star w(u,v) = \sum_{(u',v') \in S} z (u-u',v-v')\, w(u',v')~.
\end{equation}
Since $z(u,v)$ is defined in a finite domain of $(u,v)$, boundary issues
can be solved by extending $z$ with zeros or as a periodic signal. We use zero-padding extensions for the next sections, except for the last section, where we use periodic convolutions. Both cases give similar accurcies.

The network takes as input a color image
$x(u,v_0)$, or any type of multichannel signal indexed by $v_0$.  
The first layer computes a sum of convolutions of $x(u,v_0)$ along $u$, 
with filters $w_{1,v_0,v_1} (u)$
\begin{equation}
\label{nsdfi3}
x_1 (u, v_1) = \rho \Big( \sum_{v_0} x(\cdot,v_0) \star w_{1,v_0,v_1} (u) \Big)~.
\end{equation}
For any $j \geq 2$, $W_j$ computes convolutions of $x_{j-1} (u,v)$
for $v = (v_1,...,v_{j-1})$ with a family of
filters $\{ w_{v_j} \}_{v_j}$ indexed by the new attribute $v_j$:
\begin{equation}
\label{nsdfi}
x_j (u,v,v_j) = \rho \Big( x_{j-1} \star w_{v_j} (u,v) \Big)\,.
\end{equation}
As explained in \cite{mallat2016understanding},
$W_j$ has two roles. First, these convolutions indexed by $v_{j}$
prepares the discriminability 
(\ref{discrimin}) of the next layer $x_{j+1}$,
despite local or global summations along $(u,v_1,...,v_{j-1})$ implemented
at this next layer. It thus propagates discriminative variations
of $x_{j-1}$ from $(u,v_1,...,v_{j-1})$ into $v_j$. Second, each convolution with
$w_{v_j}$ computes local or global invariants by summations
along $(u,v_1,...,v_{j-2})$, in order to reduce dimensionality. 
This dimensionality reduction is implemented by 
a subsampling of $(u,v)$ at the output (\ref{nsdfi}),
which we omitted here for simplicity. 
Since $v_{k}$ is the index of multidimensional filters,
a translation along $v_{k}$ is a shift along an ordered set of
multidimensional filters. For any $k < j-1$,
$\sum_{v_{k}} x_{j-1} (u,v_1,...,v_k,...,v_{j-1})$ is invariant to any such shift.

The final operator $W_J$ computes invariants
over $u$ and all 
attributes $v_k$ but the last one:
\begin{equation}
\label{nsdifsdf}
x_{J} (v_{J-1}) = \sum_{u,v_1,...,v_{J-1}}  x_{J-1} (u,v_1,...,v_{J-1})~.
\end{equation}
The last attribute $v_{J-1}$ corresponds to the class index, and its size is 
the number of classes. The class $y = f(x)$
is estimated by applying a soft-max operator on $x_J(v_{J-1})$. 

\begin{figure*}[t]
\begin{center}
\begin{tikzpicture}[draw=black!50,y=-1cm]
\def \t{4}
\node at (0,0) [line width=0] (x0) {$x(u,v_0)$};
\node at (1.2,0) [draw=one,align=center,rounded corners=0.1cm,line width=.3] (W1) {\small $ \rho W_1$};
\node at (2.4,0) [draw=one,align=center,rounded corners=0.1cm,line width=.3] (W2) {\small $ \rho W_2$};
\node at (3.6,0) [draw=one,align=center,rounded corners=0.1cm,line width=.3] (W3) {\small $ \rho W_3$};
\node at (4.8,0) [draw=one,align=center,rounded corners=0.1cm,line width=.3] (W4) {\small $ \rho W_4$};
\node at (6,0) [draw=one,align=center,rounded corners=0.1cm,line width=.3] (W5) {\small $ \rho W_5$};
\node at (7.2,0) [draw=one,align=center,rounded corners=0.1cm,line width=.3] (W6) {\small $ \rho W_6$};
\node at (8.4,0) [draw=one,align=center,rounded corners=0.1cm,line width=.3] (W7) {\small $ \rho W_7$};
\node at (9.6,0) [draw=one,align=center,rounded corners=0.1cm,line width=.3] (W8) {\small $ \rho W_8$};
\node at (10.8,0) [draw=one,align=center,rounded corners=0.1cm,line width=.3] (W9) {\small $ \rho W_9$};
\node at (12,0) [draw=one,align=center,rounded corners=0.1cm,line width=.3] (W10) {\small $ \rho W_{10}$};
\node at (13.2,0) [draw=one,align=center,rounded corners=0.1cm,line width=.3] (W11) {\small $ \rho W_{11}$};
\node at (14.4,0) [draw=one,align=center,rounded corners=0.1cm,line width=.3] (W12) {\small $ W_{12}$};
\node at (15.6,0)  (xJ) {$x_J(v_{J-1})$};
\draw[->,thick,black] (x0) edge (W1) ;
\draw[->,thick,black] (W1) edge (W2) node[midway,sloped,left,rotate=270](e1){};
\draw[->,thick,black] (W2) edge (W3);
\draw[->,thick,black] (W3) edge (W4);
\draw[->,thick,black] (W4) edge (W5);
\draw[->,thick,black] (W5) edge (W6);
\draw[->,thick,black] (W6) edge (W7);
\draw[->,thick,black] (W7) edge (W8);
\draw[->,thick,black] (W8) edge (W9);
\draw[->,thick,black] (W9) edge (W10);
\draw[->,thick,black] (W10) edge (W11);
\draw[->,thick,black] (W11) edge (W12);
\draw[-<,thick,black] (W12) edge (xJ);
\node at (0.1,0.3) [] (s0) {\tiny$ N^2\times 3$};
\node at (0.8,1.15) [] (x1) {$ x_1(u,v_1)$};
\node at (0.9,1.5) [] (s1) {\tiny$ N^2\times K$};
\node at (2.8,1.15) [] (x2) {\small $ x_2(u,v_1,v_2)$};
\node at (2.9,1.5) [] (s2) {\tiny$ N^2\times K \times K$};
\node at (5.2,1.15) [] (x3) { $ x_3(u,v_1,v_2,v_3)$};
\node at (5.3,1.5) [] (s3) {\tiny $ N^2\times \frac{K}{4} \times \frac{K}{2} \times K$};
\node at (8.2,1.15) [] (x5) { $ x_5(u,v_3,v_4,v_5)$};
\node at (8.2,1.5) [] (s5) {\tiny $ \frac{N^2}{4}\times \frac{K}{4} \times \frac{K}{2} \times K$};
\node at (11.3,1.15) [] (x2) { $ x_9(u,v_7,v_8,v_9)$};
\node at (11.3,1.5) [] (s5) {\tiny $ \frac{N^2}{16}\times\frac{K}{4} \times \frac{K}{2} \times K$};
\node at (15.4,0.3) [] (sJ) {\tiny$ 10/100$};
\draw[black,dashed](1.8,0) -- (1,1);
\draw[black,dashed](3.0,0) -- (3,1);
\draw[black,dashed](4.15,0) -- (5.2,1);
\draw[black,dashed](6.55,0) -- (8.2,1);
\draw[black,dashed](11.3,0) -- (11.3,0.9);
\end{tikzpicture}
\end{center}
   \caption{Implementation of a multiscale hierarchical convolutional network as a cascade of $5D$ convolutions $W_j$. 
The figure gives the size of the intermediate layers
stored in $5D$ arrays. Dash dots lines indicate the parametrization of a layer $x_j$ and its dimension. We only represent dimensions when the output has a different size from the input.}
\label{fig:archi}
\end{figure*}
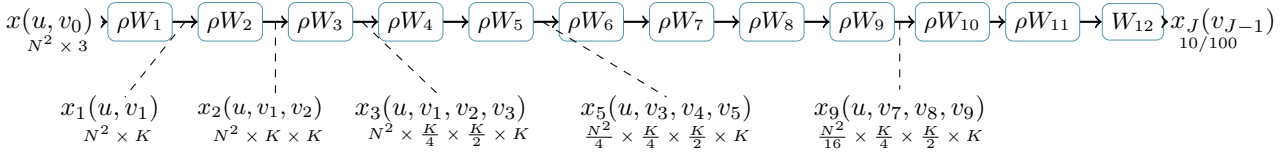

\begin{proposition}
The last layer $x_J$ is invariant to translations
of $x_{j} (u,v_1,...,v_j)$ along $(u,v_1,...,v_j)$, for any $j < J-1$. 
\end{proposition}

{\it Proof:} Observe that
$x_J = W_J\, \rho\,W_{J-1}\,...\rho\,W_j\,x_j$.
Each $W_k$ for $j < k < J$ 
is a convolution
along $(u,v_0,...,v_j,...,v_k)$ and hence covariant to translations
of $(u,v_0,...,v_j)$.
Since $\rho$ is a pointwise operator, it is also covariant to translations.
Translating $x_{j}$ along $(u,v_1,...,v_j)$ thus translates
$x_{J-1}$. Since (\ref{nsdifsdf}) computes a sum over these indices,
it is invariant to these translations.
$\Box$ \\

This proposition proves that the soft-max of $x_J$ approximates the
classification function $f_{j} (x_{j}) = f(x)$ by an operator which
is invariant to translations along the high-dimensional index
$(u,v) = (u,v_1,...,v_j)$. The change of variable $x_{j}$ thus
aims at mapping the symmetry group of $f$ into a high-dimensional
translation group, which is a flat symmetry group with no curvature. 
It means that 
classes of $x_{j}$ where $f_{j} (x_{j})$ is constant define surfaces
which are progressively more flat as $j$ increases.
However, this requires an important word of caution.
A translation of $x_j (u,v_1,...,v_j)$ along $u$ corresponds to a translation
of $x(u,v_0)$ along $u$. On the contrary, a translation along the 
attributes $(v_1,...,v_j)$ 
usually does not correspond to transformations on $x$.
Translations of $x_j$ along $(v_1,...,v_j)$ is a group of symmetries 
of $f_{j}$ but do not define transformations of $x$ and hence do not 
correspond to a symmetry group of $f$. Next sections analyze the properties of
translations along attributes computed numerically.

Let us give examples over 
images or audio signals $x(u)$ having a single channel. 
The first layer (\ref{nsdfi3}) computes convolutions along $u$:
$x_1 (u,v_1) = \rho(x \star w_{v_1} (u))$.
For audio signals, $u$ is time. This first layer usually
computes a wavelet spectrogram,
with wavelet filters $w_{v_1}$ indexed by a log-frequency index $v_1$.
A frequency transposition corresponds to a log-frequency translation 
$x_1 (u,v_1 - \tau)$ along $v_1$. If $x$ is a
sinusoidal wave then this translation corresponds to a shift of its
frequency and hence to a transformation of $x$.
However, for more general signals $x$, there exists no $x'$
such that $\rho(x' \star w_{v_1} (u)) = x_1 (u,v_1 - \tau)$. 
It is indeed well known that a
frequency transposition does not define an exact signal transformation.
Other audio attributes such as timber are not either well defined
transformations on $x$ although important attributes to classify sounds.

For images, $u = (u_1,u_2)$ is a spatial index.
If $w_{v_1} = w(r_{v_1}^{-1} u)$ is a rotation of a filter 
$w(u)$ by an angle $v_1$ then
\[
x_1 (u,v_1 - \tau) = \rho(x_\tau \star w_{v_1} (r_{\tau} u))~~
\mbox{with}~~x_\tau (u) = x(r_\tau^{-1} u). 
\]
However, there exists no $x'$ such that 
$\rho(x \star w_{v_1} (u)) = x_1 (u,v_1 - \tau)$ because of the missing 
spatial rotation $r_\tau u$. These examples show that translation $x_j (u,v_1,..,v_j)$ along the
attributes $(v_1,...,v_j)$ usually do not correspond to a transformation of $x$.

\section{Fast Low-Dimensional Architecture}
\label{fast}

\subsection{Dimensionality Reduction}

Multiscale hierarchical network layers
are indexed by two-dimensional spatial indices $u = (u_1,u_2)$ and
progressively higher dimensional attributes $v = (v_1,...,v_j)$.
To avoid computing high-dimensional vectors and convolutions, we introduce an image
classification architecture which
eliminates the dependency relatively to all attributes but the
last three $(v_{j-2},v_{j-1},v_j)$, for $j > 2$. Since $u = (u_1,u_2)$,
all layers are stored in five dimensional arrays.

The network takes as an input a color image 
$x(u,v_0)$, with three color channels $1 \leq v_0 \leq 3$ and $u = (u_1,u_2)$.
Applying (\ref{nsdfi3}) and (\ref{nsdfi}) up to $j = 3$ computes
a five-dimensional layer $x_3 (u,v_1,v_2,v_3)$. 
For $j > 3$, $x_j$ is computed 
as a linear combination of marginal sums of $x_{j-1}$
along $v_{j-3}$. Thus, it does not depend anymore
on $v_{j-3}$ and can
be stored in a five-dimensional array indexed by $(u,v_{j-2},v_{j-1},v_j)$.
This is done by convolving $x_{j-1}$ with a a filter $w_{v_j}$ which does
not depend upon $v_{j-3}$:
\begin{equation}
\label{filsdnf}
w_{v_j} (u,v_{j-3},v_{j-2},v_{j-1}) = w_{v_j} (u,v_{j-2}, v_{j-1})~.
\end{equation}
We indeed verify that this convolution 
is a linear combination of sums over $v_{j-3}$, so
$x_j$ depends only upon $(u,v_{j-2},v_{j-1},v_j)$.
The convolution is subsampled by $2^{s_j}$ with $s_j \in \{0,1\}$
along $u$, and a factor $2$ along $v_{j-1}$ and $v_j$
\[
x_j (u,v_{j-2},v_{j-1},v_j) = x_{j-1} \star w_{v_j} (2^{s_j} u, 2 v_{j-2}, 2 v_{j-1})~,
\]

At depth $j$, the array of attributes $v = (v_{j-2},v_{j-1},v_j)$ 
is of size $K/4 \times K/2 \times K$. The parameters $K$ and
spatial subsmapling factors $s_j$ are adjusted with 
a trade-off between computations, memory and classification accuracy.
The final layer is computed with a sum (\ref{nsdifsdf}) over all parameters
but the last one, which corresponds to the class index:
\begin{equation}
\label{nsdifsdf}
x_{J} (v_{J-1}) = \sum_{u,v_{J-3},v_{J-2}}  x_{J-1} (u,v_{J-3},v_{J-2},v_{J-1})~.
\end{equation}
This architecture is illustrated in Figure \ref{fig:archi}.

\subsection{Filter Factorization for Training}

Our newly introduced Multiscale Hierarchical Convolution Networks (HCNN) have been tested on 
CIFAR10 and CIFAR100 image databases. CIFAR10 has 10 classes, while CIFAR100 has 100 classes, which makes it more challenging.
The train and test sets have $50k$ and $10k$ colored images of $32\times32$ pixels.
Images are preprocessed via a standardization along the RGB channels. No whitening is applied as we did not observe any improvement. 

Our HCNN is trained in the same way as a classical CNN. We train it by minimizing a neg-log entropy loss, via SGD with momentum $0.9$ for 240 epochs. An initial learning rate of 0.25 is chosen while being reduced by a factor 10 every 40 epochs. Each minibatch is of size 50. 
The learning is regularized
by a weight decay of $2\,10^{-4}$ \cite{krizhevsky2012imagenet}. We incorporate a data augmentation with random translations of 6 pixels and flips \cite{krizhevsky2010convolutional}. 

Just as in any other CNNs, the gradient descent is badly conditioned because of the large number of parameters 
\cite{goodfellow2014qualitatively}. We precondition and regularize 
the 4 dimensional filters $w_{v_j}$, by normalizing a factorization of these filters. We factorize $w_{v_j} (u,v_{j-3},v_{j-2},v_{j-1})$ into a sum of $Q$ separable filters:
\begin{equation}
\label{factoFilters}
w_{v_j} (u,v_{j-3},v_{j-2},v_{j-1})=\sum_{q=1}^{Q} h_{j,q}(u) \,g_{v_j,q}(v_{j-2},v_{j-1}) \,, 
\end{equation}
and introduce an intermediate normalization before the sum. 
Let us write
$h_{j,q}(u,v) = \delta(u) \,h_{j,q}(u)$ and $g_{v_j,q}(u,v) = \delta(u) \,g_{v_j,q} (v)$. The batch normalization is applied to
$x_{j-1} \star h_{j,q}$ and 
substracts a mean array $m_{j,q}$ while normalizing the standard deviations
of all coefficients $\sigma_{j,q}$:
\[
\tilde x_{j,q}(u,v)=\frac{x_{j-1} \star h_{j,q}
-m_{j,q}}{\sigma_{j,q}} .
\]
This normalized output is retransformed according to (\ref{factoFilters})
by a sum over $q$ and a subsampling:
\[
x_{j}(u,v)=\rho \Big(\sum_{q=1}^Q \tilde x_{j,q} \star g_{v_j,q} (2^{s_j} u, 2 v) \Big) .
\]
The convolution operator $W_j$ is thus subdivided into a 
first operator $W^h_j$ which computes standardized convolutions along
$u$ cascaded with $W^g_j$ which sums $Q$ convolutions along $v$.
Since the tensor rank of $W_j$ cannot be larger than 9, using $Q\geq 9$ does not restrict the rank of the operators $W_j$. However, as reported in \cite{jacobsen2016structured}, increasing the value of $Q$ introduces an
overparametrization which regularizes the optimization. 
Increasing $Q$ from $9$ to $16$ and then from $16$ to $32$
brings a relative increase of the classification accuracy 
of $4.2\%$ and then of $1.1\%$.

We also report a modification of our network (denoted by (+) ) which incorporates
an intermediate non-linearity:
\[
x_{j}(u,v)=\rho (W^g_j \rho (W^h_j x_{j-1} ) )~.
\]

Observe that in this case, $x_j$ is still covariant with the actions of the translations along $(u,v)$, yet the factorization of  $w_{v_j}$ into $(h_{j,q},g_{v_j,q})$ does not hold anymore.

For classification of CIFAR images, the total depth is $J = 12$ and 
a downsampling by $2$ along $u$ is applied at depth $j=5,9$. Figure \ref{fig:archi} describes our model architecture as a cascade of $W_j$ and $\rho$, and gives the size of each layer.  
Each attribute can take at most $K=16$ values.

The number of free parameters of the original architecture
is the number of parameters of the
convolution kernels $w_{v_j}$ for $1 \leq v_j \leq K$ and $2 < j < J$,
although they are factorized
into separable filters $h_{j,q} (u)$ and $g_{v_j,q}(v_{j-2},v_{j-1})$
which involve more parameters.
The filters $w_{v_j}$ have less parameters for $j=2,3$
because they are lower-dimensional convolution kernels.
In CIFAR-10,
for $3 < j < J$, each $w_{v_j}$ has a spatial support
of size $3^2$ and a support of $7 \times 11$ along $(v_{j-2},v_{j-1})$. 
If we add the $10$ filters which output the last layer,
the resulting total number of 
network parameters is approximately $0.098M$. 
In CIFAR-100, the filters rather have a
support of $11 \times 11$ along $(v_{j-2},v_{j-1})$ but the last layer
has a size $100$ which considerable increases the 
number of parameters which is approximatively $0.25M$. 

The second implementation (+) introduces
a non-linearity $\rho$ between each separable filter, so the overall
computations can not be reduced to equivalent filters $w_{v_j}$. 
There are $Q = 32$ spatial filters $h_{j,q} (u)$ of support $3 \times 3$
and $Q\,K$  filters $g_{v_j,q} (v_{j-2},v_{j-1})$ of support $7 \times 11$. 
The total number of coefficients required to parametrize $h_{j,q},g_{v_j,q}$
is approximatively $0.34M$. 
In CIFAR-100, the number of parameters becomes $0.89M$.
The total number of parameters of the implementation (+) is thus
much bigger than the original implementation which does
not add intermediate non-linearities.
Next section compares these number of parameters with architectures that have similar numerical performances.

\section{An explicit structuration}
\label{struct}

This section shows that Multiscale Hierarchical Convolution Networks have
comparable classification accuracies on the CIFAR image dataset than
state-of-the-art architectures, with much fewer parameters.
We also investigate the properties of translations along the attributes
$v_j$ learned on CIFAR.

\subsection{Classification Performance}

We evaluate our Hierarchical CNN on CIFAR-10 (table \ref{CIFAR10}) and CIFAR-100 (table \ref{CIFAR100}) in the setting explained above. Our network achieves an error of 8.6\% on CIFAR-10, which is comparable to recent state-of-the-art architectures. On CIFAR-100 we achieve an error rate of 38\%, which is about 4\% worse than the closely related all-convolutional network baseline, but our architecture has an order of magnitude fewer parameters. 

Classification algorithms using a priori defined representations or
representations computed with unsupervised algorithms have an accuracy
which barely goes above $80\%$ on CIFAR-10 \cite{oyallon2015deep}.
On the contrary, supervised CNN have an accuracy above $90\%$ as shown
by Table \ref{CIFAR10}. This is also the case for our structured 
hierarchical network which has an accuracy above $91\%$. Improving these
results may be done with larger $K$ and $Q$ which could be done with
faster GPU implementation of multidimensional convolutions,
although it is a technical challenge \cite{budden2016deep}. Our proposed architecture is based on ``plain vanilla'' CNN architectures to which we compare
our results in Table \ref{CIFAR10}. 
Applying residual connections \cite{he2016deep}, densely connected layers \cite{huang2016densely}, or similar improvements,
might overcome the  $4\%$ accuracy gap with the best existing
architectures. In the following,
we study the properties resulting from the hierarchical structuration
of our network, compared with classical CNN.

\subsection{Reducing the number of parameters}
The structuration of a Deep neural network aims at reducing the number of parameters and making them easier to interpret in relation to signal models. Reducing the number of parameters means characterizing better 
the structures which govern the classification. 

\begin{table}
\caption{Classification accuracy on CIFAR10 dataset.}
\label{CIFAR10}
\vskip 0.15in
\begin{center}
\begin{small}
\begin{sc}
\begin{tabular}{lcccr}
\hline
\abovespace\belowspace
Model & \# Parameters & \% Accuracy \\
\hline

\abovespace
Hiearchical CNN    &0.098M & 91.43\\
\belowspace
Hiearchical CNN (+) &0.34M & 92.50\\

All-CNN  & 1.3M & 92.75\\
ResNet 20  &0.27M & 91.25\\
Network in Network  &0.98M & 91.20\\
WRN-student &0.17M &  91.23\\
FitNet  &2.5M &91.61\\
\hline
\end{tabular}
\end{sc}
\end{small}
\end{center}
\vskip -0.1in
\end{table}

This section compares multiscale hierarchical Networks
to other structured architectures and algorithms which reduce 
the number of parameters of a CNN during, and after training. 
We show that Multiscale Hierarchical Convolutional Network involves less
parameters during and after training than
other architectures in the literature.


We review various strategies to reduce the number of parameters of a CNN and compare them with our multiscale hierarchical CNN. 
Several studies show that one can factorize CNN filters
\cite{denton2014exploiting,jaderberg2014speeding} 
\textit{a posteriori}. A  reduction of parameters is obtained by
computing low-rank factorized approximations of the filters calculated by
a trained CNN. It leads to more efficient computations with operators
defined by fewer parameters.
Another strategy to reduce the number of network weights 
is to use teacher and student networks \cite{zagoruyko2016paying,romero2014fitnets}, which optimize a CNN defined by fewer parameters. The student network adapts a reduced number of parameters for data classification via the teacher network.

A parameter redundancy 
has also been observed in the final fully connected layers used 
by number of neural network architectures,
which contain most of the CNN parameters
\cite{cheng2015exploration,lu2016learning}. 
This last layer is replaced by a circulant matrix during the 
CNN training, with no loss in accuracy, which indicates that 
last layer can indeed be structured. Other approaches \cite{jacobsen2016structured} represent the filters with few parameters 
in different bases, instead of imposing tha they have a small spatial
support. These filters are represented 
as  linear combinations of a given family of filters,
for example, computed with derivatives Gaussians. This approach is 
structuring jointly the channel and spatial dimensions. Finally, HyperNetworks \cite{ha2016hypernetworks} permits to drastically reducing the number of parameters used during the training step, to $0.097M$ and obtaining  $91.98\%$ accuracy. However, we do not report them as $0.97M$ corresponds to a non-linear number of parameters for the network.

\begin{table}
\caption{Classification accuracy on CIFAR100 dataset.}
\label{CIFAR100}
\vskip 0.15in
\begin{center}
\begin{small}
\begin{sc}
\begin{tabular}{lcccr}
\hline
\abovespace\belowspace
Model & \# Parameters & \% Accuracy \\
\hline
\abovespace
Hiearchical CNN    &0.25M & 62.01\\
\belowspace
Hiearchical CNN (+) &0.89M& 63.19\\

All-CNN  &1.3M & 66.29\\
Network in Network & 0.98M & 64.32\\
FitNet  &2.5M &64.96\\
\hline
\end{tabular}
\end{sc}
\end{small}
\end{center}
\vskip -0.1in
\end{table}
Table \ref{CIFAR10} and \ref{CIFAR100} give the performance of different CNN architectures with
their number of parameters, for the CIFAR10 and CIFAR100 datasets.
For multiscale hierarchical networks, the convolution
filters are invariant to translations along $u$ and $v$ which reduces the number
of parameters by an important factor compared to other architectures. 
All-CNN \cite{springenberg2014striving} is an architecture based only on sums
of spatial convolutions and ReLU non-linearities, which has a total of $1.3M$ parameters, and a similar accuracy to ours. Its architecture is 
similar to our hierarchical architecture, but it has much more parameters
because filters are not translation invariant along $v$. Interestingly, a ResNet \cite{he2016deep} has more parameters and performs similarly whereas it is a more complex architecture, due to the shortcut connexions. WRN-student is a student resnet \cite{zagoruyko2016paying} with 0.2M parameters trained via a teacher using 0.6M parameters and which gets an accuracy of $93.42\%$ on CIFAR10. FitNet networks \cite{romero2014fitnets} also use compression methods but need at least 2.5M parameters, which is much larger than our network. Our architecture brings an important
parameter reduction on CIFAR10 for accuracies around $90\%$ There is also a drastic reduction of parameters on CIFAR100. 

\subsection{Interpreting the translation}

\begin{figure}[t]
\center
\includegraphics[width=8cm]{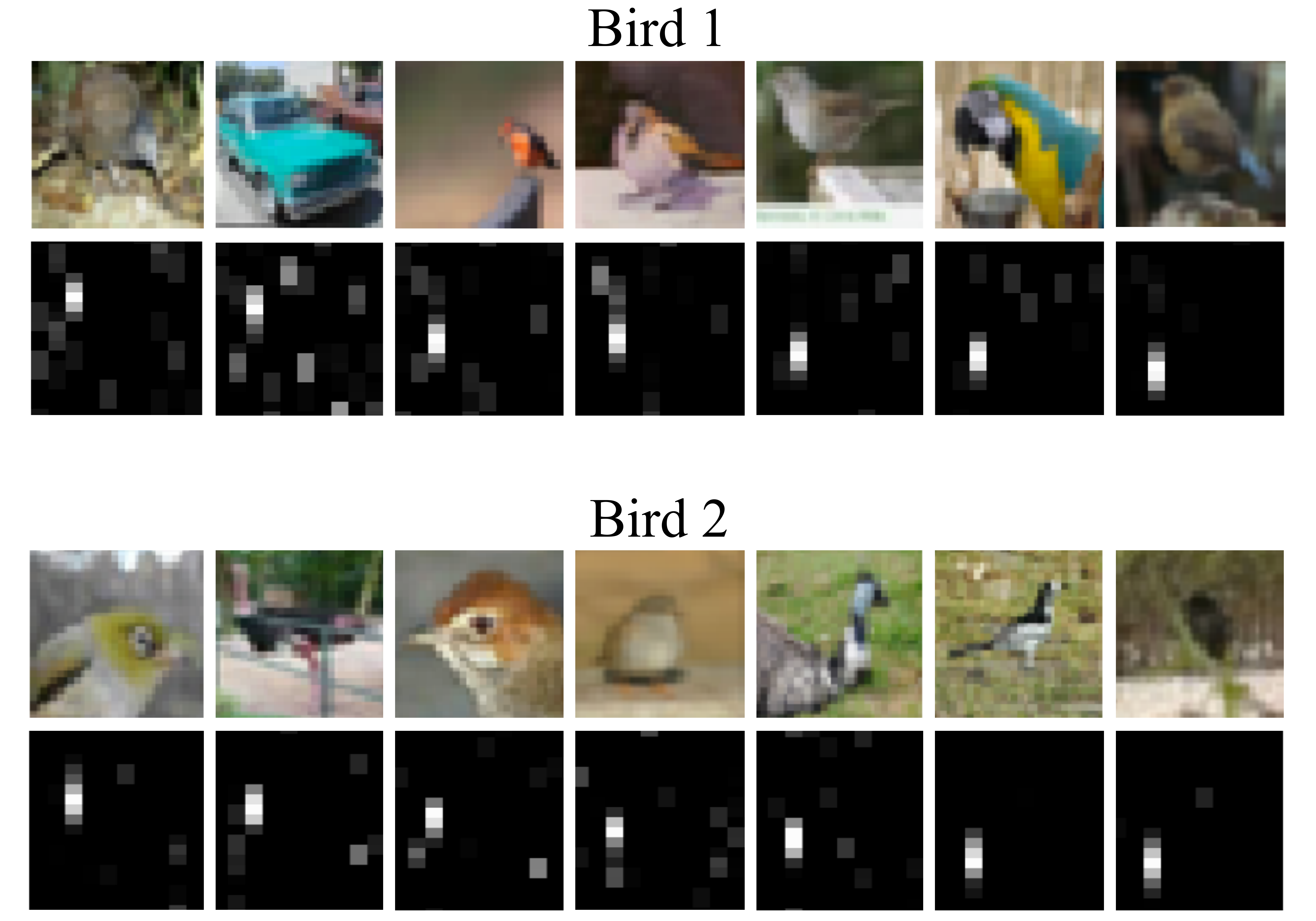}
\caption{The first images of the first and third rows are the two input
image $x$. Their invariant attribute array
$\bar x_{j} (v_{j-1},v_{j})$ is shown below for $j = J-1$, with high
amplitude coefficients appearing as white points.
Vertical  and horizontal axes correspond respectively to
$v_{j-1}$ and $v_j$, so translations of $v_{j-1}$ by $\tau$ 
are vertical translations. An image $x^\tau$ in a column $\tau+1$ has
an invariant attribute $\bar x^\tau_j$ which is shown below. It
is the closest to $\bar x_j (v_{j-1}-\tau,v_j)$ in the databasis.}
\label{fig:figure_translation}
\end{figure}

The structure of Multiscale Hierarchical CNN opens up the possibility 
of interpreting inner network coefficients, which is usually not
possible for CNNs. 
A major mathematical challenge is to understand the type of invariants
computed by deeper layers of a CNN. 
Hierarchical networks
computes invariants to translations relatively
to learned attributes $v_j$, which are indices of the filters $w_{v_j}$.
One can try to relate  these attributes
translations to modifications of image properties.
As explained in Section \ref{MultiHiera}, a translation of
$x_j$ along $v_{j}$ usually does 
not correspond to a well-defined 
transformation of the input signal $x$ but it produces
a translation of the next layers. Translating $x_j$ along
$v_j$ by $\tau$ 
translates $x_{j+1} (u,v_{j-1},v_j,v_{j+1})$ along $v_j$ by $\tau$.

\begin{figure*}[t]
\center
\includegraphics[width=10cm]{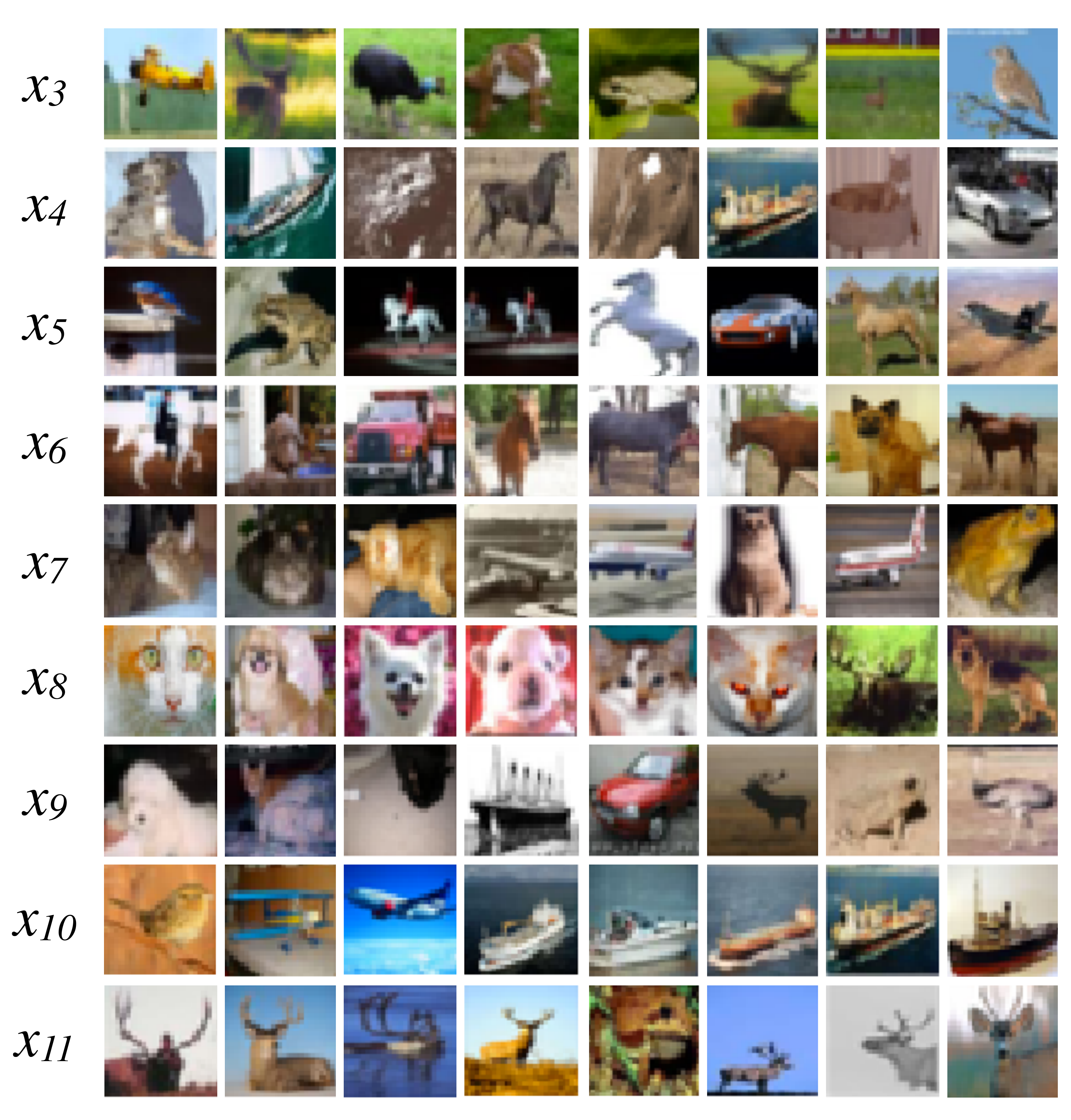}
\caption{The first columns give the input image $x$, from which
we compute the invariant array $\bar x_j$ at a depth $3 \leq j \leq 11$ 
which increases with the row.
The next images in the same row are the images $x^\tau$ whose 
invariant arrays $\bar x^\tau_j$ are the closest to $\bar x_j$ translated
by $1 \leq \tau \leq 7$, among all other images in the databasis. 
The value of $\tau$ is the column index minus $1$.}
\label{fig:figure_different_depth}
\end{figure*}

To analyze the effect of this translation, we eliminate variability along $v_{j-2}$ and define an invariant attribute array
by choosing the central spatial position $u_0$:
\begin{equation}
\label{invdsndfsdf}
\bar x_j(v_{j-1},v_j)=\sum_{v_{j-2}} x_j(u_0,v_{j-2},v_{j-1},v_{j}).
\end{equation}
We relate this translation to an image  in the training dataset
by finding the image  $x^\tau$ in the dataset which minimizes
$\|\bar x_j(v_{j-1}-\tau,v_j) - \bar x_j^\tau (v_{j-1},v_j)\|_2$, 
if this minimum Euclidean distance is sufficiently small. 
To compute accurately a translation by $\tau$ 
we eliminate the high frequency
variations of $x_j$ and $x^\tau_j$ along  $v_{j-1}$ with a  filter which
averages consecutive samples, before computing their translation.
The network used in this experiment is implemented with circular convolutions to avoid border effects, which have nearly the same classification
performance. 

Figure \ref{fig:figure_translation} shows the sequence of 
$x^\tau$ obtained with a translation by $\tau$ of $\bar x_j$ at depth $j=J-1$, for two images $x$ in the ``bird'' class. Since we are close
to the ouptut, we expect that
translated images belong to the same class. This is not the
case for the second image of 
the first "Bird 1". It is a "car" instead of a "bird".
This corresponds to 
a classification error but observe that $\bar x^\tau_{J-1}$
is quite different from $\bar x_{J-1}$ translated.
We otherwise  observe that in these final layers, translations of
$\bar x_{J-1}$ defines images in the same class.

Figure \ref{fig:figure_different_depth} gives sequences
of translated attribute images $x^\tau$, computed
by translating $\bar x_j$ by $\tau$ at different depth $j$ and 
for different input $x$. 
As expected, at small depth $j$, translating an attribute $v_{j-1}$ does
not define images in the same class. These attribute rather correspond
to low-level image properties which depend upon fine scale image
properties. However, these low-level properties can
not be identified just by looking at these images. Indeed, 
the closer images $x^\tau$ identified in the databasis are obtained with
a distance over coefficients which are
invariant relatively to all other attributes. These images are thus
very different and involve variabilities relatively to all other
attributes. 
To identify the nature of an attribute $v_j$, 
a possible approach is to correlate the images $x^\tau$
over a large set of images, while modifying known properties
of $x$. 

At deep layers $j$, 
translations of $\bar x_j$ 
define images $x^r$ which have a progressively higher probability
to belong to the same class as $x$. 
These attribute transformations correspond to large scale image
pattern related 
to modifications of the filters $w_{v_{j-1}}$.
In this case, the attribute indices could be interpreted
as addresses in organized arrays. 
The translation group would then correspond to translations 
of addresses. Understanding better the properties of attributes at
different depth is an issue that will be explored in the future. 

\section{Conclusion}
Multiscale Hierarchical convolutional networks give a mathematical
framework to study invariants computed by deep neural networks. 
Layers are parameterized
in progressively higher dimensional spaces of hierarchical
attributes, which are learned
from training data. All network
operators are multidimensional convolutions along attribute indices,
so that invariants can be computed by summations along these
attributes.

This paper gives image classification results with an efficient
implementation computed with a cascade of 
5D convolutions and intermediate non-linearities. Invariant
are progressively calculated as depth increases.
Good classification accuracies are obtained with
a reduced number of parameters compared to other CNN.

Translations along attributes at shallow depth correspond to low-level
image properties at fine scales whereas attributes at deep layers
correspond to modifications of large scale pattern structures.  
Understanding better the multiscale properties of these attributes
and their relations to the symmetry group of $f$ is an important issue,
which can lead to a better mathematical understanding
of CNN learning algorithms.

\section*{Acknowledgements} 
This work is funded by STW project ImaGene, ERC grant InvariantClass 320959 and via a grant for
PhD Students of the Conseil r\'{e}gional d'\^{Il}e-de-France (RDMIdF).

\pagebreak
\bibliography{example_paper}
\bibliographystyle{icml2017}

\end{document}